\documentclass[mathpazo]{jmlc-OTH}

\usepackage{hyperref}
\usepackage{url}
\usepackage[capitalize]{cleveref}
\usepackage{overpic}
\usepackage{xcolor}
\usepackage{graphicx}
\usepackage{algorithm}
\usepackage{algpseudocode}

\usepackage{enumitem}
\usepackage{booktabs}    

\usepackage{jabbrv}
\DefineSpuriousJournalWord{on}
\DefineSpuriousJournalWord{Its}
\DefineJournalAbbreviation{National}{Natl} 

\newcommand{\mb}[1]{{\mathbf{#1}}}

\begin{document}
\title{Deep Embedded Multiplicative DMD\\ for Algebra-Preserving Koopman Learning}


\author[Gray K et.~al.]{Kelan Gray\corrauth\affil{1},
    Finlay Brown\affil{1}, Nicolas Boull\'e\affil{1}, and Matthew J. Colbrook \affil{2}}
\address{\affilnum{1}\ Department of Mathematics,
    Imperial College London,
    London, SW7 2AZ, UK. \\
    \affilnum{2}\ Department of Applied Mathematics and Theoretical Physics,
    University of Cambridge, Cambridge, CB3 0WA, UK.}
\emails{{\tt k.gray25@imperial.ac.uk}, {\tt f.brown24@imperial.ac.uk}, {\tt n.boulle@imperial.ac.uk}, {\tt m.colbrook@damtp.cam.ac.uk}.}

\begin{abstract}
    Koopman theory turns nonlinear dynamics into a linear spectral problem. In computation, however, everything depends on a hard finite-dimensional choice: the observables must be expressive, nearly invariant under the dynamics, and, ideally, compatible with composition. Deep Koopman methods learn flexible coordinates, whereas structure-preserving methods enforce operator identities on fixed dictionaries. We combine these ideas by introducing Deep Embedded Multiplicative Dynamic Mode Decomposition (DeepMDMD), a method that learns a latent space and a partition of it, while enforcing the Koopman product rule as an exact algebraic constraint. Training alternates between an exact multiplicative operator update and a differentiable latent-clustering step that promotes Koopman closure. The result is a finite transition map on learned latent cells. Its nonzero spectrum lies on the unit circle, its dictionary is shaped by the dynamics rather than by ambient geometry, and forecasts are made in latent coordinates before being decoded to physical space. Across Hamiltonian, chaotic, and fluid examples, DeepMDMD learns dictionaries that are far more compact and dynamically coherent than those produced by geometric MDMD partitions. It reduces spectral pollution, reveals richer continuous-spectrum structure, and gives stable forecasts under severe noise. In high-dimensional flows, including a 158,624-dimensional cylinder wake and a noisy $\mathrm{Re}=20,000$ lid-driven cavity, it preserves coherent structures and long-time spectral statistics where state-space MDMD fails.
    These results suggest a practical rule for Koopman learning: learn the coordinates, constrain the algebra.
\end{abstract}


\maketitle

\section{Introduction} \label{Intro}

We consider discrete-time dynamical systems
\begin{align}
    \mb{x}_{n+1} = \mb{F}(\mb{x}_n), \qquad n \ge 0,
    \label{dynamical system}
\end{align}
where $\mb{x}_0\in\mathcal X \subset \mathbb{R}^d$ is the state space and
$\mb{F}: \mathcal X \rightarrow \mathcal X$ is the evolution map.
In fluid mechanics \cite{schmid2010dynamic}, climate dynamics
\cite{froyland2021spectral}, neuroscience \cite{brunton2016extracting},
and many other areas, $d$ is large and the dynamics are nonlinear. Direct
analysis of \eqref{dynamical system} is then often very difficult.
A data-driven approach assumes instead a collection of snapshot pairs
\begin{equation}
    \{(\mb{x}^{(m)}, \ \mb{y}^{(m)} = \mb{F}(\mb{x}^{(m)}))\}_{m=1}^M,
    \label{snapshots}
\end{equation}
and seeks to infer the dynamics from these data.
The Koopman viewpoint \cite{koopman1931hamiltonian, koopman1932dynamical}
moves the problem from states to observables, that is, functions of the state.
The reward is linearity: observables evolve under a linear Koopman operator.
The price is dimension: this operator is generally infinite-dimensional. Its
spectrum encodes the dynamics. Eigenvalues give growth rates and frequencies,
while eigenfunctions reveal coherent structures. Thus numerical methods that
approximate Koopman spectra accurately are central to data-driven dynamical
systems \cite{mezic2004comparison, mezic2005spectral,brunton2021modern, budivsic2012applied, colbrook2024multiverse,
    klus2015numerical}.

\begin{figure}[t]
    \centering
    \includegraphics[width=1\textwidth]{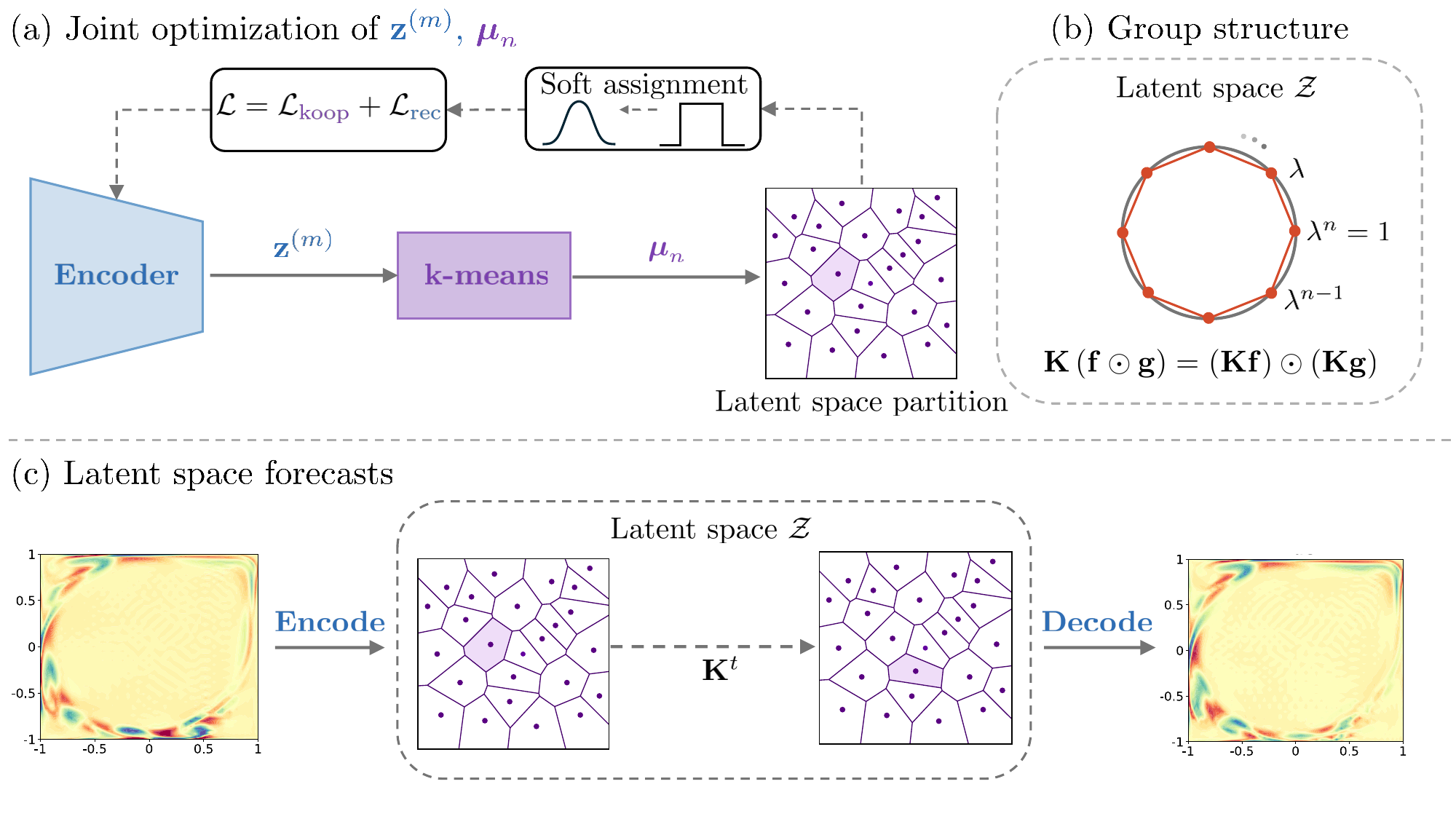}
    \vspace{-1cm}
    \caption{Schematic of the DeepMDMD pipeline: (a) the latent encodings $\mb z^{(m)}$ and cluster centroids $\boldsymbol \mu_n$ are jointly optimized to produce a dynamically coherent partition of the latent space $\mathcal{Z}$; (b) the DeepMDMD algorithm yields a Koopman approximation supported on $\mathcal{Z}$ that enforces the multiplicative structure as a hard constraint; (c) forecasting is performed entirely in $\mathcal{Z}$, with predictions decoded to the full state space $\mathcal{X}$ only at output time.}
    \label{fig:pipeline}
\end{figure}

The standard computational tool is \textit{Extended Dynamic Mode Decomposition}
(EDMD) \cite{williams2015data}, a finite-dimensional Galerkin approximation
of the Koopman operator. A central difficulty is the choice of dictionary. The chosen observables should span a space that is expressive, low-dimensional, and nearly closed under the Koopman action. These requirements are hard to reconcile.
Structure-preserving variants of DMD
\cite{baddoo2023physics, colbrook2023mpedmd, drmavc2024hermitian,
    boulle2025convergence, lu2020lagrangian, salova2019koopman, morandin2023port,
    krake2022constrained, cohen2020mode, huang2018data, boulle2025multiplicative} build known properties of the Koopman operator into the finite-dimensional
problem. This often leads to better approximations and more efficient use of data.
In this paper we focus on \textit{Multiplicative Dynamic Mode
    Decomposition} (MDMD) \cite{boulle2025multiplicative}. The Koopman operator
acts by composition, and therefore satisfies the product rule:
$\mathcal K(fg) = (\mathcal K f)(\mathcal K g)$, which is enforced by MDMD as a constraint. This naturally leads to
dictionaries of indicator functions. The resulting finite-dimensional
operators respect the multiplicative structure of the Koopman operator and
substantially reduce spectral pollution.
For MDMD, dictionary selection becomes a partitioning problem. One
must divide the state space into cells whose indicator functions form the
dictionary. The partition is no minor detail; it determines the approximation.
Existing implementations typically use $k$-means. This is simple and robust,
but it is geometric rather than dynamical. It sees distances, not evolution.
Consequently, accurate approximation may require large dictionaries. In
high-dimensional systems the problem is worse: $k$-means suffers from the curse
of dimensionality \cite{bishop2006pattern}, and the resulting clusters, hence
basis functions, may be poorly aligned with the dynamics.

We propose \textit{Deep Embedded Multiplicative Dynamic Mode Decomposition}
(DeepMDMD) to replace this geometry-first step by a learned one. Instead of
partitioning the state space $\mathcal X$, we partition the latent space of a
pretrained autoencoder \cite{goodfellow2016deep}. The latent representation and
the clustering are then refined jointly to promote Koopman closure. The result
is a compact, interpretable dictionary adapted to the dynamics, while
multiplicativity of the Koopman approximation remains a hard algebraic
constraint. Forecasts are computed in latent space and decoded only when
physical-space output is needed. This makes long rollouts far cheaper than
evolving in the full state space $\mathcal X$. A schematic of the method is
shown in \cref{fig:pipeline}.

The rest of the paper is organized as follows. In \cref{prelim} we review the
necessary preliminaries and related work. In \cref{method} we introduce
DeepMDMD, including its joint training strategy and latent-space forecasting
procedure. In \cref{experiments} we test spectral accuracy, forecasting, flow
statistics, and robustness to noise. We conclude in \cref{conclusion}.

\section{Preliminaries and related work} \label{prelim}

\subsection{The Koopman operator}

Let $\omega$ be a positive measure on $\mathcal X$, and suppose that
composition with $\mb F$ is well defined on $L^2(\mathcal X,\omega)$. The
Koopman operator is
\begin{equation*}
    \mathcal K : L^2(\mathcal X,\omega) \to L^2(\mathcal X,\omega),
    \qquad
    [\mathcal K g](\mb x) = g(\mb F(\mb x)).
\end{equation*}
Thus the nonlinear dynamics in \cref{dynamical system} induce a linear,
generally infinite-dimensional, evolution of observables. This is the basis
for spectral approaches to nonlinear dynamics
\cite{mezic2005spectral, mezic2020spectrum, korda2020data,
    colbrook2024rigorous, colbrook2025introductory}.

A key property of $\mathcal K$ is multiplicativity
\cite{boulle2025multiplicative}. Whenever $f,g\in L^2(\mathcal X,\omega)$ such that $fg \in L^2(\mathcal X,\omega)$,
\begin{equation}\label{mult property}
    [\mathcal K(fg)](\mb x)
    =
    [\mathcal K f](\mb x)\,[\mathcal K g](\mb x),
    \qquad \mb x \in \mathcal X .
\end{equation}
Thus $\mathcal K$ preserves pointwise products wherever those products belong
to the observable space. In particular, if $\lambda_1$ and $\lambda_2$ are
Koopman eigenvalues with eigenfunctions $\phi_1$ and $\phi_2$, then, whenever
the product is admissible,
\begin{equation} \label{closed_eigs}
    [\mathcal K(\phi_1^n\phi_2^m)](\mb x)
    =
    \lambda_1^n \lambda_2^m
    \phi_1^n(\mb x)\phi_2^m(\mb x),
    \qquad n,m\in\mathbb N .
\end{equation}
This multiplicative closure is the algebraic source of the group structure
seen in Koopman spectra.

Throughout the paper we assume that $\mathcal K$ is unitary. Equivalently,
the map $\mb F$ is invertible and measure preserving with respect to $\omega$
\cite[Chap.~7]{eisner2015operator}. Hence the spectrum lies on the unit
circle $\mathbb T=\{z\in\mathbb C: |z|=1\}$, and both the spectrum and the
point spectrum are unions of subgroups of $\mathbb T$
\cite{ridge1973spectrum}. This setting includes many systems of interest:
Hamiltonian dynamics \cite{arnold1989mathematical}, equilibrium physical
systems \cite{hill2012introduction}, ergodic systems
\cite{walters2000introduction}, and systems whose long-time behavior is
well described by measure-preserving dynamics after transients
\cite{mezic2005spectral}.

\subsection{Extended Dynamic Mode Decomposition}\label{EDMD}

Let $\{\psi_1,\dots,\psi_N\}\subset L^2(\mathcal X,\omega)$ be a fixed
dictionary, and set $\mathcal V_N=\operatorname{span}\{\psi_1,\dots,\psi_N\}$. Extended Dynamic Mode Decomposition (EDMD) uses the snapshot pairs
\eqref{snapshots} to approximate the compression of the Koopman operator to
$\mathcal V_N$ \cite{williams2015data}. Define the row-vector feature map
\[
    \mb\Psi(\mb x)
    =
    [\psi_1(\mb x),\dots,\psi_N(\mb x)] \in \mathbb C^{1\times N},
\]
so that any $g\in\mathcal V_N$ can be written as
$g(\mb x)=\mb\Psi(\mb x)\mb g$ for some coefficient vector
$\mb g\in\mathbb C^N$. EDMD seeks a matrix
$\mb K_{\textup{EDMD}}\in\mathbb C^{N\times N}$ such that
\[
    [\mathcal K g](\mb x)
    =
    \mb\Psi(\mb F(\mb x))\mb g
    \approx
    \mb\Psi(\mb x)\mb K_{\textup{EDMD}}\mb g,
    \qquad \mb g\in\mathbb C^N.
\]
From the data, form the matrices
\begin{equation*}
    \mb{\Psi}_X =
    \begin{pmatrix}
        \mb \Psi(\mb{x}^{(1)}) \\
        \vdots                 \\
        \mb \Psi(\mb{x}^{(M)})
    \end{pmatrix}\in \mathbb{C}^{M\times N},
    \qquad
    \mb{\Psi}_Y =
    \begin{pmatrix}
        \mb \Psi(\mb{y}^{(1)}) \\
        \vdots                 \\
        \mb \Psi(\mb{y}^{(M)})
    \end{pmatrix} \in \mathbb{C}^{M\times N}.
\end{equation*}
To approximate the $L^2(\mathcal X,\omega)$ inner product, we assign positive
quadrature weights $w_m>0$ to the data points and write
$\mb W=\operatorname{diag}(w_1,\dots,w_M)$. The weighted EDMD problem is
\begin{equation}\label{EDMD problem}
    \min_{\mb K \in \mathbb{C}^{N \times N}}
    \left\|
    \mb W^{1/2}
    \left(\mb{\Psi}_Y-\mb{\Psi}_X\mb K\right)
    \mb C^{-1}
    \right\|_{\textup F}^2 ,
\end{equation}
where the nonsingular matrix $\mb C\in\mathbb C^{N\times N}$ specifies the
norm in which the residual is measured. For the least-squares minimizer itself
this choice is immaterial: the Moore--Penrose solution is
\[
    \mb K_{\textup{EDMD}}
    =
    \left(\mb \Psi_X^* \mb W \mb \Psi_X\right)^\dagger
    \mb \Psi_X^* \mb W \mb \Psi_Y .
\]
The matrix $\mb C$ will, however, play an important role in the constructions
below. For fixed $N$, and under a convergent quadrature rule,
$\mb K_{\textup{EDMD}}$ converges as $M\to\infty$ to the Galerkin matrix of
$\mathcal P_{\mathcal V_N}\mathcal K|_{\mathcal V_N}$, equivalently of
$\mathcal P_{\mathcal V_N}\mathcal K\mathcal P_{\mathcal V_N}$ on
$L^2(\mathcal X,\omega)$ \cite{korda2018convergence}.

The effectiveness of EDMD depends strongly on the choice of
$\mathcal V_N$. Spectral information is captured faithfully only when this
space is close to invariant under the Koopman operator; exact invariance is
rare in practice. Designing such dictionaries is especially difficult for
complex or high-dimensional systems. A common example is a dictionary of radial
basis functions, which requires choosing $N$ centers in $\mathcal X$. If $n$
centers are placed in each coordinate direction on a tensor-product grid, then
$N=n^d$, exposing EDMD to the curse of dimensionality
\cite{li2017extended,williams2015data}. Neural networks offer another route beyond fixed dictionaries. In EDMD with
dictionary learning \cite{li2017extended}, the least-squares problem
\eqref{EDMD problem} is optimized over both the Koopman matrix and a
parametrized dictionary. Related methods learn nonlinear embeddings into latent
coordinates in which the dynamics are approximately linear
\cite{takeishi2017learning, yeung2019learning, lusch2018deep,
    otto2019linearly, azencot2020forecasting, mardt2018vampnets}. Typically an encoder
maps the state to a latent representation, which is then advanced by a linear
operator. Finally, recent works aimed to approximate the Koopman operator on an RKHS and used kernel feature maps as dictionary that can be optimized through a posteriori error estimates \cite{boulle2025convergent,conradie2026trustworthy}.

These methods can be powerful, but their flexibility is also a weakness. The
learned coordinates are often largely unconstrained and need not respect the
algebraic or physical structure of the Koopman operator. This can obscure
interpretation and impair generalization. Our approach aims to retain the expressive power of a learned latent space while imposing the multiplicative structure used by MDMD.

\subsection{Koopman mode decomposition}

Let $\mb K\in\mathbb C^{N\times N}$ be a finite-dimensional Koopman
approximation, and suppose that it is diagonalizable:
\[
    \mb K=\mb V\mb\Lambda\mb V^{-1},
    \qquad
    \mb\Lambda=\operatorname{diag}(\lambda_1,\dots,\lambda_N).
\]
The columns $\mb v_j$ of $\mb V$ define approximate Koopman eigenfunctions
\[
    \varphi_j(\mb x)=\mb\Psi(\mb x)\mb v_j .
\]
Thus any observable $g\in\mathcal V_N$, written as
$g(\mb x)=\mb\Psi(\mb x)\mb g$, has the expansion
\[
    g(\mb x)
    =
    \mb\Psi(\mb x)\mb V\mb\Xi,
    \qquad
    \mb\Xi=\mb V^{-1}\mb g .
\]
The coefficients $\mb\Xi$ are the Koopman-mode coefficients of $g$ in this
eigenfunction basis. If $g\notin\mathcal V_N$, one first replaces $g$ by its
least-squares, or Galerkin, approximation in $\mathcal V_N$.

This representation gives the Koopman mode decomposition
\cite{mezic2005spectral}. Along a trajectory,
\begin{equation} \label{KMD 1}
    g(\mb x_n)
    =
    [\mathcal K^n g](\mb x_0)
    \approx
    \mb\Psi(\mb x_0)\mb K^n\mb g
    =
    \mb\Psi(\mb x_0)\mb V\mb\Lambda^n\mb\Xi .
\end{equation}
The evolution is therefore separated into spatial structures
$\varphi_j$, temporal factors $\lambda_j^n$, and mode coefficients.

A central example is the full-state observable $g:\mb x\mapsto \mb x$.
Applying \cref{KMD 1} componentwise gives the state forecast
\[
    \mb x_n
    \approx
    \mb\Psi(\mb x_0)\mb V\mb\Lambda^n\boldsymbol{\Xi},
\]
where the Koopman modes
$\boldsymbol{\Xi}\in\mathbb C^{N\times d}$ are vectors in physical space.
They are obtained from the data matrix
\[
    \mb X=
    [\mb x^{(1)},\dots,\mb x^{(M)}]^\top
    \in\mathbb R^{M\times d}
\]
by the least-squares fit
\[
    \boldsymbol{\Xi}
    =
    (\mb\Psi_X\mb V)^\dagger \mb X .
\]
Once a factorization of $\mb\Psi_X\mb V$ is available, forming this fit costs
$\mathcal O(MNd)$; the factorization itself adds the usual least-squares cost.
Since $N$ often grows rapidly with the state dimension $d$, full-state KMD can
become expensive in high-dimensional problems. It is also sensitive to noise:
perturbations in the snapshots can destabilize the computed spectral data and
therefore the forecast \cite{bagheri2014effects}.

\subsection{Multiplicative Dynamic Mode Decomposition}

Multiplicative Dynamic Mode Decomposition (MDMD) constructs finite-dimensional
Koopman approximations that preserve the product rule
\eqref{mult property} \cite{boulle2025multiplicative}. This requires an
observable space closed under pointwise multiplication. A natural choice is
given by indicator functions on a measurable partition. For
$\{S_n\}_{n=1}^N\in\Sigma_N(\mathcal X)$, define
\begin{equation*}
    \psi_n(\mb x) =
    \begin{cases}
        1, & \mb x \in S_n,    \\
        0, & \text{otherwise}.
    \end{cases}
\end{equation*}
These functions satisfy $\psi_i\psi_j=\delta_{ij}\psi_i$, and hence their
span is closed under multiplication. The admissible partitions are
\begin{equation} \label{partition space}
    \Sigma_N(\mathcal X)
    =
    \left\{
    \{S_n\}_{n=1}^N \subset \mathcal B(\mathcal X):
    S_n\cap S_{n'}=\emptyset \ \text{for } n\ne n',
    \quad
    \bigcup_{n=1}^N S_n=\mathcal X
    \right\},
\end{equation}
where $\mathcal B(\mathcal X)$ denotes the Borel $\sigma$-algebra.

For this dictionary the feature vector is
\[
    \mb\Psi(\mb x)
    =
    [\psi_1(\mb x),\ldots,\psi_N(\mb x)]\in\mathbb C^{1\times N}.
\]
If $f,g\in\mathcal V_N$ have coefficient vectors $\mb f,\mb g\in\mathbb C^N$,
then the disjointness of the supports gives
\[
    f(\mb x)g(\mb x)
    =
    \mb\Psi(\mb x)(\mb f\odot \mb g),
    \qquad \mb x\in\mathcal X,
\]
where $\odot$ denotes componentwise multiplication. MDMD therefore seeks a
matrix $\mb K_{\textup{MDMD}}\in\mathbb C^{N\times N}$ satisfying
\begin{equation}  \label{mult constraint}
    \mb K_{\textup{MDMD}}(\mb f\odot\mb g)
    =
    (\mb K_{\textup{MDMD}}\mb f)
    \odot
    (\mb K_{\textup{MDMD}}\mb g),
    \qquad \mb f,\mb g\in\mathbb C^N .
\end{equation}

The empirical inner product on $\mathcal V_N$ is represented by the Gram matrix
\[
    \mb G=\mb\Psi_X^*\mb W\mb\Psi_X .
\]
Since the supports are disjoint, $\mb G$ is diagonal:
\[
    \mb G=\operatorname{diag}(G_1,\dots,G_N),
    \qquad
    G_n=\sum_{\mb x^{(m)}\in S_n} w_m .
\]
We assume here that each cluster has positive empirical weight, so that
$\mb G^{-1/2}$ is well defined.

For a fixed partition $\{S_n\}_{n=1}^N\in\Sigma_N(\mathcal X)$, and with
$\mb C=\mb G^{1/2}$ in \cref{EDMD problem}, MDMD solves
\begin{equation} \label{mult problem}
    \min_{\substack{\mb K\in\mathbb C^{N\times N}\\
            \mb K\ \text{satisfies \cref{mult constraint}}}}
    \left\|
    \mb W^{1/2}
    \left(\mb\Psi_Y-\mb\Psi_X\mb K\right)
    \mb G^{-1/2}
    \right\|_{\textup F}^2 .
\end{equation}
The constraint \eqref{mult constraint} is highly restrictive. It forces
$\mb K_{\textup{MDMD}}$ to have entries in $\{0,1\}$ and at most one nonzero
entry in each row. Thus $\mb K_{\textup{MDMD}}$ defines a finite transition
map on the clusters: a row with a one in column $j$ sends the corresponding
cluster to $S_j$, while a zero row represents a terminating state. Iteration of
this finite map must eventually enter a cycle or terminate, decomposing the
partition into cyclic and transient components.

Consequently, the nonzero eigenvalues of $\mb K_{\textup{MDMD}}$ are roots of
unity. They lie on the unit circle, while the remaining eigenvalues are zero,
so MDMD rules out exponentially growing spectral components in the KMD rollout
\eqref{KMD 1}. For an indicator dictionary, the minimizer of
\cref{mult problem} can be computed in $\mathcal O(M+N^2)$ operations; see
\cite{boulle2025multiplicative} for details.

The partition $\{S_n\}_{n=1}^N\in\Sigma_N(\mathcal X)$ is often chosen by
$k$-means clustering \cite{lloyd1982least}. This places centroids
$\{\boldsymbol\mu_n\}_{n=1}^N\subset\mathcal X$ and induces a Vorono\"i
tessellation of the state space \cite{aurenhammer1991voronoi}. The difficulty
is that $k$-means is geometric, not dynamical. It favors compact Vorono\"i
cells rather than sets adapted to the action of $\mb F$, and the resulting
partition boundaries need not respect the dynamics
\cite[Fig. 4]{boulle2025multiplicative}. The corresponding indicator
dictionary may then have poor closure under the Koopman operator, so accurate
approximations can require many clusters.

This difficulty is amplified in high dimensions, where $k$-means deteriorates
under the curse of dimensionality. A common remedy is to first apply proper
orthogonal decomposition (POD) and then cluster the reduced coordinates. But
POD is a linear projection, and it may discard nonlinear features that are
important for systems with complex or chaotic dynamics.

\section{Methodology} \label{method}

A key bottleneck in applying MDMD to high-dimensional systems is the choice of
partition. One wants sets $\{S_n\}_{n=1}^N$ with good closure properties under
the dynamics, but constructing such a partition directly in
$\mathcal X\subset\mathbb R^d$ is difficult when $d$ is large. DeepMDMD
addresses this difficulty by learning a low-dimensional latent space
$\mathcal Z\subset\mathbb R^k$, with $k\ll d$, and placing the MDMD partition
in $\mathcal Z$ rather than in $\mathcal X$.

Let $\mb z^{(m)}$ and $\widetilde{\mb z}^{(m)}$ denote the latent embeddings of
$\mb x^{(m)}$ and $\mb y^{(m)}$, respectively. For a partition
$\{S_n\}_{n=1}^N\in\Sigma_N(\mathcal Z)$, let
$\mb\Psi_Z$ and $\mb\Psi_{\widetilde Z}$ denote the corresponding indicator
dictionary evaluated at the current and forward latent snapshots. We seek both
the partition and the finite-dimensional Koopman approximation by minimizing
\begin{equation}
    \min_{\substack{
    \mb K \in \mathbb{C}^{N \times N},\,
    \{S_n\}_{n=1}^N \in \Sigma_N(\mathcal Z)\\
    \mb K \text{ satisfies \cref{mult constraint}}
    }}
    \left\|
    \mb W^{1/2}
    \left(
    \mb\Psi_{\widetilde Z}
    -
    \mb\Psi_Z \mb K
    \right)
    \mb G^{-1/2}
    \right\|_{\mathrm F}^2 .
    \label{joint problem}
\end{equation}

The latent space is initialized with an autoencoder pretrained for
reconstruction, and the initial partition is obtained by $k$-means in this
space. Since reconstruction alone is not dynamical, these initial cells need
not be coherent under the map. We therefore optimize \cref{joint problem} by
alternating between two steps:
\begin{enumerate}[leftmargin=*]
    \item \textit{Operator update}. With the partition fixed, compute
          $\mb K$ exactly by the MDMD algorithm. This preserves the multiplicative
          constraint.

    \item \textit{Partition update}. With $\mb K$ fixed, update the latent
          embeddings and centroids $\{\boldsymbol\mu_n\}_{n=1}^N$ by gradient
          descent. Because hard cluster assignments are nondifferentiable, we use a
          soft relaxation based on the Student $t$-kernel
          \cite{maaten2008visualizing}, and restore hard assignments at the next
          operator update.
\end{enumerate}

The result is a Koopman approximation acting on latent indicator functions.
The partition is low-dimensional, hence less exposed to the curse of
dimensionality, and it is shaped by the dynamics rather than fixed purely by
geometry. After training, forecasts are computed in latent space and decoded
only when physical-space output is required, making rollouts substantially
cheaper than evolving in $\mathcal X$.

\begin{algorithm}[t]
    \caption{The DeepMDMD algorithm}\label{alg:deepmdmd}
    \begin{algorithmic}[1]
        \State \textbf{Input:} Snapshots $\{(\mathbf{x}^{(m)}, \mathbf{y}^{(m)})\}_{m=1}^{M}$, weights $\mathbf W$,
        clusters $N$, $\lambda \ge 0$, $T_{\mathrm{update}}$
        \State Pretrain $\mathbf{E}_\theta$, $\mathbf{D}_\theta$ by minimizing $\mathcal{L}_{\mathrm{rec}}$
        \State Encode latent pairs $\{\mathbf{z}^{(m)}, \tilde{\mathbf{z}}^{(m)}\}$
        \State Initialize $\{\boldsymbol{\mu}_n\}_{n=1}^{N}$ via $k$-means$\texttt{++}$
        \Repeat
        \State Build dictionary matrices $\boldsymbol{\Psi}_Z$, $\boldsymbol{\Psi}_{\smash{\widetilde{Z}}}$ on latent space from $\{\boldsymbol{\mu}_n\}_{n=1}^{N}$
        \State Compute  Gram matrix $\mb G$ via \cref{latent gram} and  $\mathbf{K} = \mathrm{MDMD}(\boldsymbol{\Psi}_Z, \boldsymbol{\Psi}_{\smash{\widetilde{Z}}}, \mb W)$ with \cref{mult problem}
        \For{$s = 1, \ldots, T_{\mathrm{update}}$}
        \State Form soft assignments $\mathbf{Q}_Z$, $\mathbf{Q}_{\smash{\widetilde{Z}}}$ via Student's $t$-kernel \cref{student t-kernel}
        \State Update $\theta$ and $\{\boldsymbol{\mu}_n\}_{n=1}^{N}$ by minimizing
        $\mathcal{L} = \mathcal{L}_{\mathrm{koop}} + \lambda\,\mathcal{L}_{\mathrm{rec}}$
        \EndFor
        \Until{convergence}
        \State \textbf{Output:} $\mathbf{E}_\theta$, $\mathbf{D}_\theta$, $\{\boldsymbol{\mu}_n\}_{n=1}^{N}$, $\mathbf{K}$
    \end{algorithmic}
\end{algorithm}

\subsection{Autoencoder pretraining}

We first construct a latent space $\mathcal Z\subset\mathbb R^k$, with
$k\ll d$, by training an autoencoder on the snapshot data \eqref{snapshots}.
The encoder $\mb E_\theta:\mathcal X\to\mathcal Z$ and decoder
$\mb D_\theta:\mathcal Z\to\mathcal X$, with parameters $\theta$, are chosen
so that $\mb D_\theta\circ \mb E_\theta$ approximates the identity on the data.
Specifically, we minimize the weighted reconstruction loss
\begin{equation*}
    \mathcal L_{\mathrm{rec}}(\theta)
    =
    \sum_{m=1}^M
    w_m
    \left\|
    \mb x^{(m)}
    -
    \mb D_\theta\left(\mb E_\theta(\mb x^{(m)})\right)
    \right\|^2 .
\end{equation*}
Here $w_m>0$ are the quadrature weights introduced in \cref{EDMD}, used to
approximate integration with respect to $\omega$.

After pretraining, we encode the snapshot pairs to obtain
\begin{equation}
    \{(\mb z^{(m)},\,\widetilde{\mb z}^{(m)})\}_{m=1}^M
    =
    \{(
    \mb E_\theta(\mb x^{(m)}),
    \mb E_\theta(\mb y^{(m)})
    )\}_{m=1}^M .
    \label{latent snaps}
\end{equation}
The initial partition $\{S_n\}_{n=1}^N\in\Sigma_N(\mathcal Z)$ is then
obtained by applying $k$-means to the latent states
$\{\mb z^{(m)}\}_{m=1}^M$. This yields centroids
$\{\boldsymbol\mu_n\}_{n=1}^N\subset\mathbb R^k$ and the corresponding
Vorono\"i cells in $\mathcal Z$.

This initialization is geometric rather than dynamical. The loss
$\mathcal L_{\mathrm{rec}}$ ignores the pairing
$\mb x^{(m)}\mapsto \mb y^{(m)}$, and hence gives the encoder no reason to
align the cells with the induced latent dynamics. The resulting indicator
dictionary may therefore have poor closure properties and may degrade the
spectral approximation. The alternating operator and partition updates below
address this by adapting the latent space and its partition to the dynamics.

\subsection{Operator update}

Given the current latent pairs
$\{(\mb z^{(m)},\widetilde{\mb z}^{(m)})\}_{m=1}^M$ and partition
$\{S_n\}_{n=1}^N\in\Sigma_N(\mathcal Z)$, we compute the Koopman matrix
$\mb K\in\mathbb C^{N\times N}$ by MDMD. Each latent point is assigned to its
Vorono\"i cell, giving the indicator dictionary
\begin{equation}
    \psi_n(\mb z) =
    \begin{cases}
        1, & \mb z \in S_n,    \\
        0, & \text{otherwise},
    \end{cases}
    \qquad n=1,\ldots,N .
    \label{hard latent clusters}
\end{equation}
Although these cells are polyhedral in $\mathcal Z$, the corresponding
functions on the original state space are $\psi_n\circ\mb E_\theta$, whose
supports $\mb E_\theta^{-1}(S_n)$ are generally nonlinear and non-polyhedral.
Thus the encoder shapes the effective indicator functions in $\mathcal X$, and
the subsequent partition update can move these sets by changing both the latent
coordinates and the centroids.

With
\[
    \mb\Psi(\mb z)
    =
    [\psi_1(\mb z),\ldots,\psi_N(\mb z)]
    \in\mathbb C^{1\times N},
\]
we form
\begin{equation}
    \mb{\Psi}_Z =
    \begin{pmatrix}
        \mb \Psi(\mb{z}^{(1)}) \\
        \vdots                 \\
        \mb \Psi(\mb{z}^{(M)})
    \end{pmatrix}\in \mathbb{C}^{M\times N},
    \qquad
    \mb{\Psi}_{\widetilde Z} =
    \begin{pmatrix}
        \mb \Psi(\widetilde{\mb{z}}^{(1)}) \\
        \vdots                             \\
        \mb \Psi(\widetilde{\mb{z}}^{(M)})
    \end{pmatrix} \in \mathbb{C}^{M\times N}.
    \label{latent dict}
\end{equation}
The empirical Gram matrix is diagonal,
\begin{equation}
    \mb G
    =
    \operatorname{diag}(G_1,\dots,G_N),
    \qquad
    G_n
    =
    \sum_{\mb z^{(m)}\in S_n} w_m ,
    \label{latent gram}
\end{equation}
where $G_n$ is the empirical mass of the $n$th latent cell. We assume that all
cells have positive mass, so that $\mb G^{-1/2}$ is well defined.

With $\mb\Psi_Z$, $\mb\Psi_{\widetilde Z}$, and $\mb G$ fixed, the operator
update is the MDMD solution of \cref{joint problem}. The resulting matrix
$\mb K$ satisfies the multiplicative constraint exactly and defines the
cluster-to-cluster transition map for the latent dynamics.

\begin{figure}[t]
    \centering
    \includegraphics[width=0.5\linewidth]{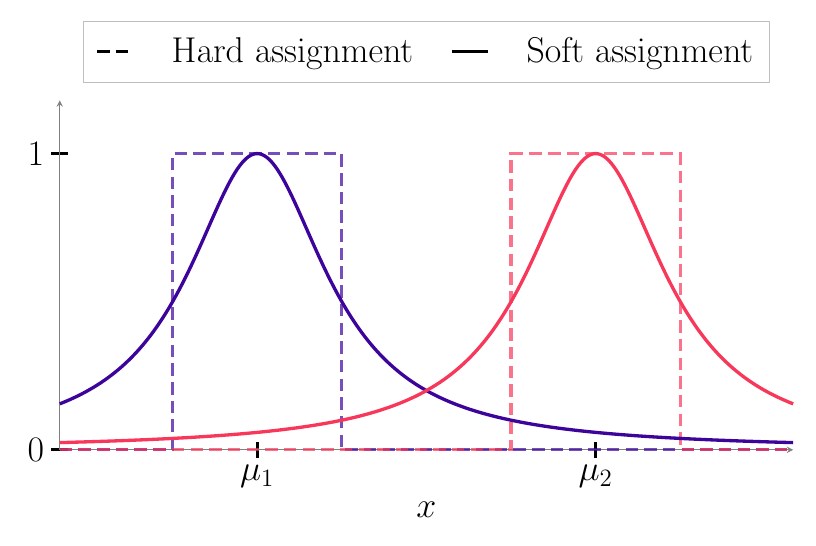}
    \caption{Comparison of hard and soft cell assignments in the latent space $\mathcal{Z}$. Hard assignments \eqref{hard latent clusters} are indicator functions supported on disjoint cells of the partition $\{S_n\}_{n=1}^N \in \Sigma_N(\mathcal{Z})$. Soft assignments are induced by the Student's $t$-kernel in  \cref{student t-kernel}, and relax the discrete partition constraint by allowing each embedding to contribute to multiple cells, enabling gradient-based optimization of $\mathbf{E}_\theta$ and $\{\boldsymbol{\mu}_n\}_{n=1}^N$.}
    \label{fig:student-t}
\end{figure}

\subsection{Partition update}

With $\mb K$ fixed, we update the encoder $\mb E_\theta$ and the centroids
$\{\boldsymbol\mu_n\}_{n=1}^N$ to improve closure of the latent partition
under the dynamics. Direct minimization of \cref{joint problem} over
$\Sigma_N(\mathcal Z)$ is not practical: the hard assignments in
\cref{hard latent clusters} are discrete, and hence nondifferentiable with
respect to both $\theta$ and the centroids. We therefore replace them by soft
assignments defined by a Student $t$-kernel \cite{maaten2008visualizing},
\begin{equation}
    [\mb Q]_{mn}
    =
    \frac{
        \left(
        1+\|\mb z^{(m)}-\boldsymbol\mu_n\|^2/\alpha
        \right)^{-(\alpha+1)/2}
    }
    {
        \sum_{n'=1}^N
        \left(
        1+\|\mb z^{(m)}-\boldsymbol\mu_{n'}\|^2/\alpha
        \right)^{-(\alpha+1)/2}
    },
    \label{student t-kernel}
\end{equation}
where $[\mb Q]_{mn}$ is the soft membership of $\mb z^{(m)}$ in cluster $n$,
and $\alpha>0$ is the degrees-of-freedom parameter. We use $\alpha=1$, giving
the Cauchy kernel. Its heavy tails make the assignments less local than those
from a Gaussian kernel, helping to alleviate the crowding effects common in
low-dimensional embeddings \cite{maaten2008visualizing}. Hard and soft
assignments are illustrated in \cref{fig:student-t}.

Replacing the indicator matrices in \cref{joint problem} by their soft
counterparts, and holding $\mb K$ and $\mb G$ fixed, gives the
Koopman-based clustering loss
\begin{equation}
    \mathcal L_{\mathrm{koop}}
    \bigl(\theta,\{\boldsymbol\mu_n\}_{n=1}^N\bigr)
    =
    \left\|
    \mb W^{1/2}
    \left(
    \mb Q_{\widetilde Z}
    -
    \mb Q_Z \mb K
    \right)
    \mb G^{-1/2}
    \right\|_{\mathrm F}^2 .
    \label{koop loss}
\end{equation}
Here $\mb Q_Z,\mb Q_{\widetilde Z}\in\mathbb R^{M\times N}$ are the soft
assignment matrices, defined by \cref{student t-kernel} and evaluated at the
latent snapshot pairs \eqref{latent snaps}. The factor $\mb G^{-1/2}$
normalizes the residual by the empirical masses of the clusters. Minimizing
\cref{koop loss} over the encoder parameters and centroids encourages a latent
partition for which forward soft assignments are predicted by the finite
Koopman transition matrix $\mb K$.

The fine-tuning objective is
\begin{equation}
    \mathcal L\bigl(\theta,\{\boldsymbol\mu_n\}_{n=1}^N\bigr)
    =
    \mathcal L_{\mathrm{koop}}
    \bigl(\theta,\{\boldsymbol\mu_n\}_{n=1}^N\bigr)
    +
    \lambda\,\mathcal L_{\mathrm{rec}}(\theta),
    \qquad \lambda>0 .
    \label{fine-tuning loss}
\end{equation}
The reconstruction term keeps the latent variables decodable and regularizes
the geometry of the embedding \cite{guo2017improved}. Unlike approaches that
train directly on multi-step prediction error \cite{lusch2018deep}, this
objective does not fit rollouts explicitly. It instead shapes the latent space
so that cluster memberships evolve according to the multiplicative Koopman
model, while preserving enough information for reconstruction and forecasting.

Choosing $\lambda$ is nontrivial. As in other deep clustering methods
\cite{xie2016unsupervised, ren2024deep, van2009learning}, there is no
ground-truth partition against which to tune it; in the present setting, an
ideal partition would be one adapted to the Koopman dynamics. We therefore use
fixed values throughout: $\lambda=0$ in low-dimensional experiments, where
reconstruction is unnecessary, and $\lambda=0.25$ in high-dimensional
experiments, where it gives stable behavior of both loss terms; see
\cref{fig:loss_history}.

We optimize $\theta$ and $\{\boldsymbol\mu_n\}_{n=1}^N$ using Adam
\cite{kingma2014adam}. The updated embeddings and centroids are then passed to
the next operator update, where the hard assignments in
\cref{hard latent clusters} are restored. To reduce cost, this update is
performed only every $T_{\mathrm{update}}$ gradient steps, allowing the latent
geometry to adjust before recomputing $\mb K$. Empirically, the alternating
scheme converges stably in our experiments; a theoretical analysis is not
pursued here. The full procedure is summarized in \cref{alg:deepmdmd}.

\begin{algorithm}[t]
    \caption{DeepMDMD Forecasting}
    \label{alg:DeepMDMD_forecast}
    \begin{algorithmic}[1]
        \State {\bfseries Input:} Koopman operator $\mb K$, dictionary $\mb \Psi_Z$, data $\{\mb x^{(m)}\}_{m=1}^M$, encoder $\mb E_\theta$, decoder $\mb D_\theta$, time horizon $T$
        \State Encode and stack: $\mathbf{Z} = [\mathbf{E}_\theta(\mathbf{x}^{(1)}), \dots, \mathbf{E}_\theta(\mathbf{x}^{(M)})]^\top \in \mathbb{R}^{M \times k}$
        \State Compute eigendecomposition $\mb K = \mb V\mb \Lambda \mb V^{-1}$
        \State Compute latent modes $\mb \Xi = (\mb \Psi_Z\mb V)^{\dagger}\mb Z$
        \For{$t = 1, \ldots, T$}
        \State Evolve latents $\hat{\mb z}_{M+t} = \mb \Psi (\mb z_{M})\mb V\mb \Lambda^t\mb \Xi$
        \State Decode $\hat{\mb x}_{M+t} = \mb D_\theta(\hat{\mb z}_{M+t})$
        \EndFor
        \State {\bfseries Output:} $\{\hat{\mb x}_{M+t}\}_{t=1}^{T}$
    \end{algorithmic}
\end{algorithm}

\subsection{Latent-space forecasting}

After training, forecasts are performed in the latent space rather than in the
$d$-dimensional state space. We use the latent full-state observable
$g:\mb z\mapsto\mb z$. Suppose that the learned Koopman matrix
$\mb K\in\mathbb C^{N\times N}$ is diagonalizable,
\[
    \mb K=\mb V\mb\Lambda\mb V^{-1},
    \qquad
    \mb\Lambda=\operatorname{diag}(\lambda_1,\dots,\lambda_N),
\]
where $\mb V$ contains the right eigenvectors. Let
\[
    \mb Z=
    [\mb z^{(1)},\dots,\mb z^{(M)}]^\top
    \in\mathbb R^{M\times k}
\]
be the matrix of latent codes. The latent Koopman modes are
\[
    \mb\Xi
    =
    (\mb\Psi_Z\mb V)^\dagger \mb Z
    \in\mathbb C^{N\times k}.
\]
These modes live in the latent space, replacing the physical-space modes in
$\mathbb C^{N\times d}$. Once a factorization of $\mb\Psi_Z\mb V$ is
available, forming the modes costs $\mathcal O(MNk)$ rather than
$\mathcal O(MNd)$. In addition, the partition is constructed in
$\mathcal Z\subset\mathbb R^k$, so the dictionary size is governed by the
latent dimension rather than the ambient dimension $d$. This mitigates the
curse of dimensionality when $k\ll d$.

The latent representation also provides a degree of robustness. In full-state
KMD, measurement noise may perturb all $d$ coordinates and hence the computed
modes. Here the forecast is restricted to the learned latent representation,
which can attenuate noise components not captured by the encoder.

For a rollout initialized at a latent state $\mb z_0$, the KMD forecast is
\begin{equation}
    \hat{\mb z}_{t}
    =
    \mb\Psi(\mb z_0)\mb V\mb\Lambda^t\mb\Xi,
    \qquad t\ge 0 .
    \label{KMD 2}
\end{equation}
Here $\mb\Psi(\mb z_0)$ encodes the cluster assignment of the initial latent
state. When forecasting from the last observed state, we take
$\mb z_0=\mb E_\theta(\mb x_M)$. Since the decoder is retained during
fine-tuning through \cref{fine-tuning loss}, latent forecasts can be mapped
back to physical space by
$\hat{\mb x}_t
    =
    \mb D_\theta(\hat{\mb z}_t).$
The full forecasting procedure is summarized in
\cref{alg:DeepMDMD_forecast}.

\section{Numerical Experiments} \label{experiments}

\begin{table}[t]
    \centering
    \small
    \setlength{\tabcolsep}{6pt}
    \caption{Training hyperparameters for each experiment.}
    \label{tab:hyperparams}
    \begin{tabular}{lcccc}
        \toprule
        \textbf{Hyperparameter}
         & \textbf{Pendulum}
         & \textbf{Lorenz 96}
         & \textbf{Cylinder Wake}
         & \textbf{Lid-Driven Cavity}                                     \\
        \midrule
        Dict.\ size $N$
         & 100/1000                   & 1000      & 80        & 500       \\
        State-space dim.\ $d$
         & 2                          & 9         & 158,624   & 4225      \\
        Latent dim.\ $k$
         & 10                         & 3         & 3         & 3         \\
        Architecture
         & $[d,128,64,k]$
         & $[d,128,64,k]$
         & $[d,256,128,64,k]$
         & $[d,256,128,64,k]$                                             \\
        Activation
         & \texttt{tanh}
         & \texttt{tanh}
         & \texttt{relu}
         & \texttt{tanh}                                                  \\
        Decoder weight $\lambda$
         & 0                          & 0.25      & 0.25      & 0.25      \\
        Pretrain epochs
         & 20                         & 250       & 500       & 500       \\
        Fine-tune epochs
         & 20                         & 100       & 250       & 250       \\
        Pretrain LR
         & $10^{-3}$                  & $10^{-3}$ & $10^{-3}$ & $10^{-3}$ \\
        Fine-tune LR
         & $10^{-3}$                  & $10^{-3}$ & $10^{-4}$ & $10^{-4}$ \\
        Dropout
         & 0                          & 0         & 0         & 0.1       \\
        \bottomrule
    \end{tabular}
\end{table}

\begin{figure}[t]
    \centering
    \includegraphics[width=1\linewidth]{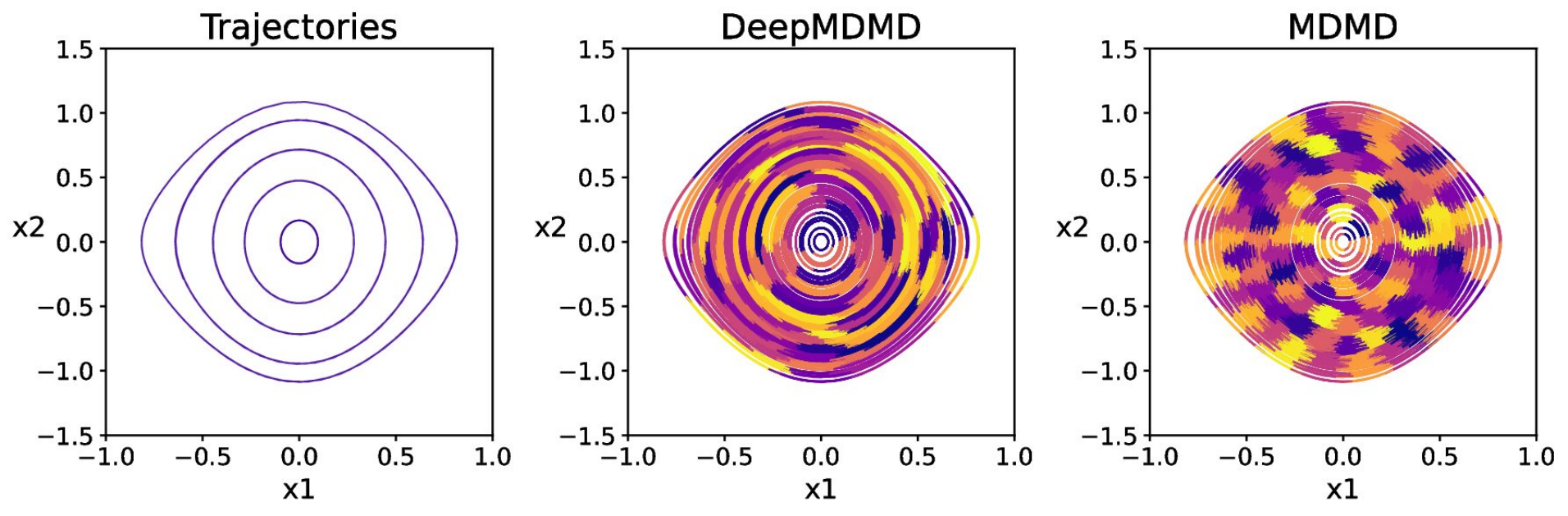}
    \caption{The nonlinear pendulum system in \cref{pendulum eqn}. Left: Example trajectories. Middle: DeepMDMD optimized partition. Right: MDMD partition initialized using $k$-means. The dictionary size is set to $N=100$.}
    \label{fig:pen clusters}
\end{figure}

We compare DeepMDMD with MDMD, the natural multiplicative baseline, on four
systems: the nonlinear pendulum, used as a low-dimensional benchmark; the
Lorenz 96 system; and two high-dimensional fluid examples, the cylinder wake
and the lid-driven cavity, both considered in the noisy regime.

\textit{Implementation details.}\quad
Unless otherwise stated, the quadrature weights are uniform,
$w_m=1/M$. DeepMDMD is implemented in PyTorch \cite{paszke2019pytorch} using
symmetric encoder--decoder architectures. Latent centroids are initialized by
the $k$-means++ algorithm \cite{arthur2007k} using scikit-learn
\cite{pedregosa2011scikit}. Operator updates are performed every
$T_{\mathrm{update}}=20$ gradient steps, and all experiments use batch size
$256$. Experiment-specific hyperparameters are listed in
\cref{tab:hyperparams}. Code and data for reproducing the numerical results
are available at
\url{https://github.com/kelangray/DeepMDMD}.

\begin{figure}[t]
    \centering
    \includegraphics[width=0.65\linewidth]{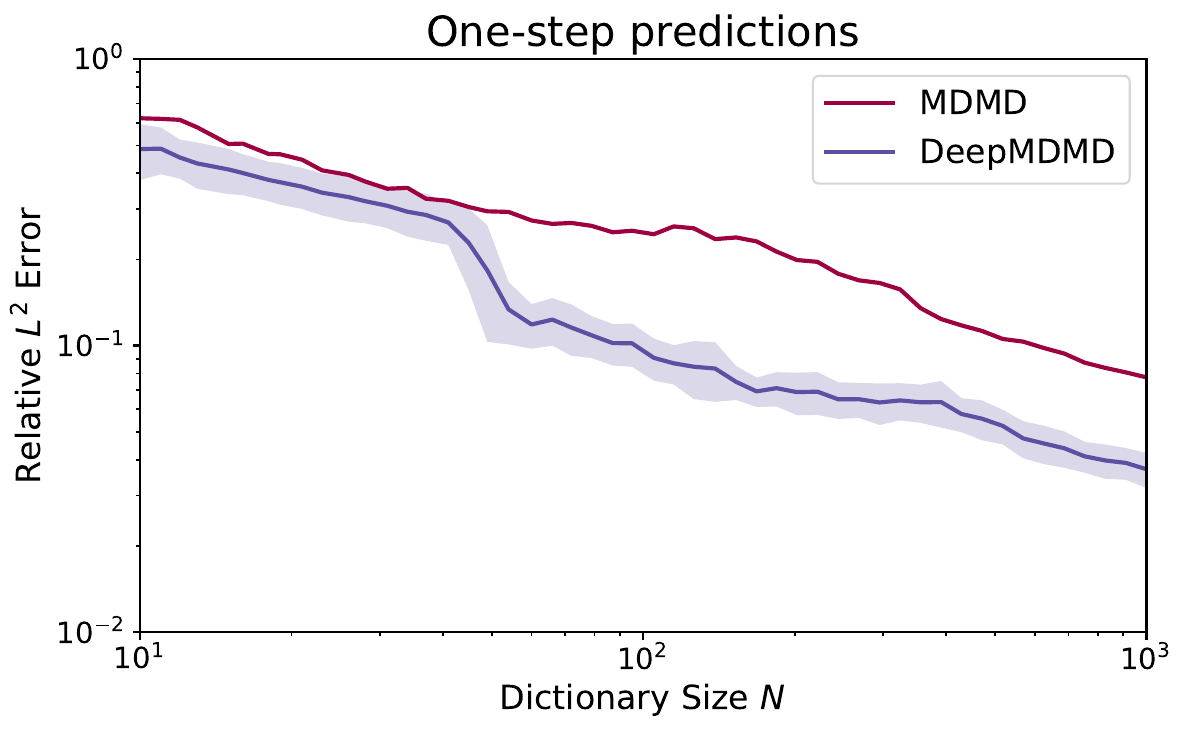}
    \caption{Nonlinear pendulum. relative $L^2$ error of one-step predictions for the Hamiltonian $h$ as a function of dictionary size $N$, comparing MDMD and DeepMDMD averaged over $50$ random seeds.}
    \label{fig:convergence pendulum}
\end{figure}

\subsection{Nonlinear pendulum}

We begin with the nonlinear pendulum, a two-dimensional test problem used here
to assess partition optimization rather than dimension reduction. The state
$\mb x=(x_1,x_2)$ satisfies
\begin{equation}
    \dot x_1 = x_2,
    \qquad
    \dot x_2 = -\sin(3x_1),
    \qquad
    (x_1,x_2)\in\mathcal X
    =
    [-\pi/3,\pi/3]_{\mathrm{per}}\times\mathbb R .
    \label{pendulum eqn}
\end{equation}
We draw $400$ initial conditions uniformly from the box $[-0.6,0.6]^2$, for
which the resulting trajectories remain in $[-1,1]^2$ over the integration
window. The system is advanced by a fourth-order Runge--Kutta method with
time step $\Delta t=0.1$ up to final time $10$, producing
$M=4\times 10^4$ snapshot pairs for the time-$\Delta t$ map associated with
\eqref{pendulum eqn}.

The Hamiltonian
\begin{equation*}
    h(x_1,x_2)
    =
    \frac12 x_2^2-\frac13\cos(3x_1)
\end{equation*}
is conserved along trajectories. Hence $h$ is a Koopman eigenfunction with
eigenvalue $\lambda=1$. The remaining nontrivial spectral content is
continuous on the unit circle, reflecting the energy-dependent oscillation
frequencies of the pendulum.

\begin{figure}[th!]
    \centering
    \begin{overpic}[width=0.9\linewidth]{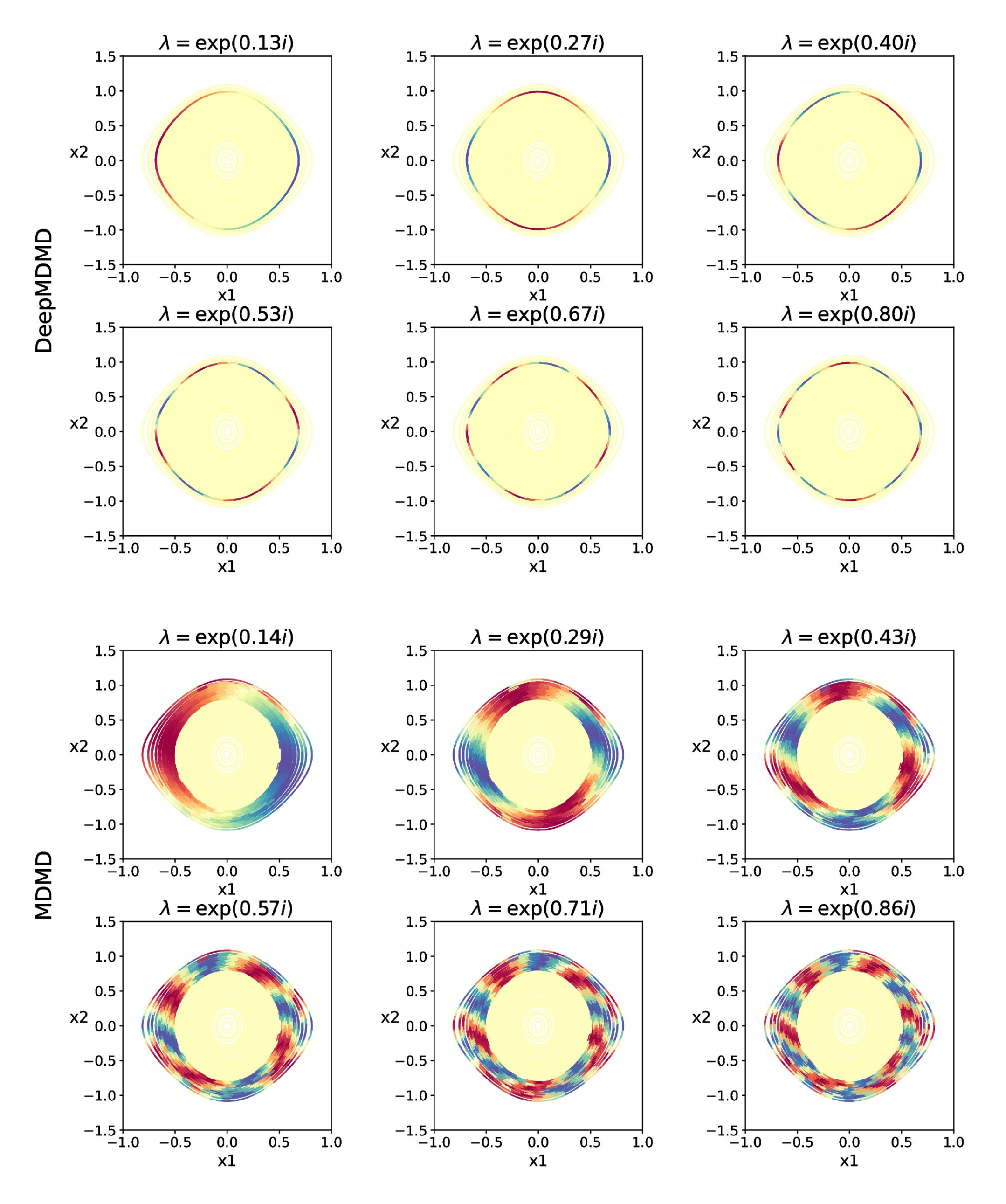}
        \put(2,95){(a)}
        \put(2,46){(b)}
    \end{overpic}
    \caption{Eigenfunctions for the pendulum system along with the corresponding eigenvalues calculated using (a) DeepMDMD and (b) MDMD.}
    \label{fig:eigfuns pendulum}
\end{figure}

We apply DeepMDMD with dictionary size $N=100$ and latent dimension $k=10$.
Since this example is already low-dimensional, we set $\lambda=0$ during
fine-tuning, removing the reconstruction term and isolating the effect of
partition optimization. The results are shown in \cref{fig:pen clusters}. The
right panel shows the $k$-means++ partition used to construct the MDMD
dictionary. Its boundaries cut across trajectories, giving an indicator space
with poor closure under the time-$\Delta t$ map. By contrast, the DeepMDMD
partition in the middle panel aligns with the trajectories and localizes the
basis functions on coherent regions of phase space. This improves both closure
and interpretability.

To quantify this effect, we compute the empirical relative $L^2$ error
\[
    \frac{
    \|\mathcal K h-\mb\Psi\mb K\mb h\|_{L^2}
    }{
    \|\mathcal K h\|_{L^2}
    },
\]
estimated by quadrature over the snapshot data. Here $\mb h$ denotes the
coefficient vector obtained by projecting the Hamiltonian $h$ onto the
indicator dictionary. Since $h$ is conserved, this error measures how well the
finite-dimensional Koopman approximation preserves the invariant Hamiltonian.
In \cref{fig:convergence pendulum} we plot the error against dictionary size
$N$, with DeepMDMD averaged over $50$ random seeds. Compared with MDMD using
$k$-means++ partitions, DeepMDMD achieves comparable accuracy with roughly an
order of magnitude fewer basis functions.

\begin{figure}[t]
    \centering
    \includegraphics[width=1\linewidth]{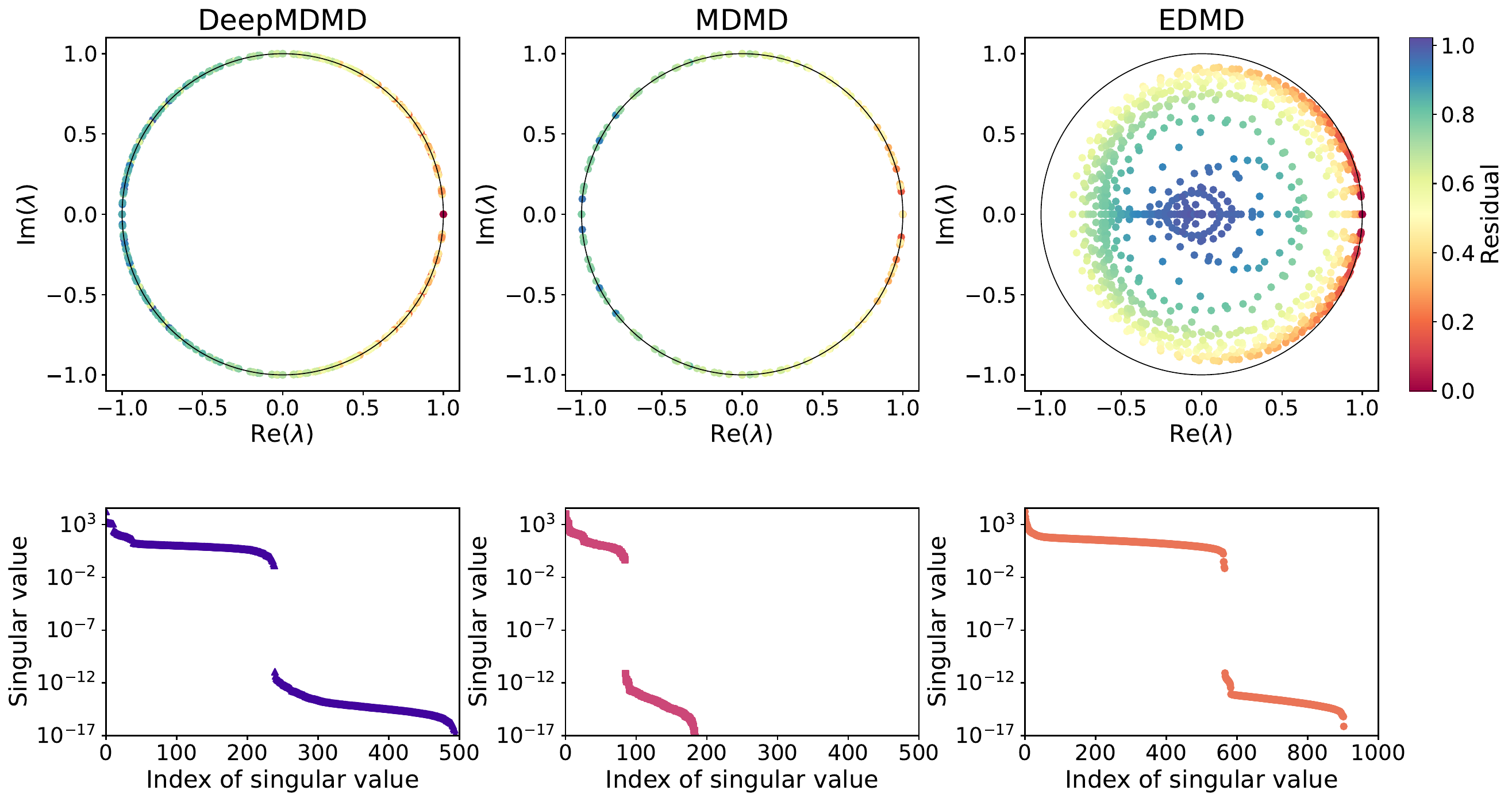}
    \caption{Top row: eigenvalues of the finite-dimensional Koopman approximation for the pendulum system \eqref{pendulum eqn}, computed using DeepMDMD (left), MDMD (middle), and EDMD with the DeepMDMD dictionary (right), colored by residual with zero eigenvalues omitted. The unit circle (black) indicates the true spectrum of $\mathcal{K}$. Bottom row: singular values of the matrix $\mb R$ for DeepMDMD (left), MDMD (middle) and EDMD (right), quantifying the number of principal eigenfunctions resolved by each method.}
    \label{fig:pendulum spectra}
\end{figure}

We next examine the spectral content of DeepMDMD. In
\cref{fig:eigfuns pendulum} we show representative eigenfunctions associated
with a finite cyclic subgroup of $\mathbb T$, computed by DeepMDMD and by MDMD
with $N=1000$ basis functions. For the nonlinear pendulum, the $L^2$ point
spectrum contains no eigenvalues other than $\lambda=1$; the nontrivial
spectral content is continuous. Nevertheless, finite-dimensional methods
produce modes that may approximate generalized eigenfunctions. For this
integrable Hamiltonian system, these generalized eigenfunctions are oscillatory
modes supported on invariant energy levels
\cite{colbrook2025rigged,mezic2020spectrum}. DeepMDMD captures this geometry:
its eigenfunctions are localized along Hamiltonian level sets. By contrast,
the MDMD eigenfunctions leak across energy levels.

We also compare the eigenvalues computed by DeepMDMD, MDMD, and EDMD, taking
the EDMD dictionary to be the same as that learned by DeepMDMD. To distinguish
reliable spectral information from pollution, we compute the empirical residual
of each approximate eigenpair $(\lambda,\mb g)$. Let
\[
    \mb A = \mb\Psi_{\widetilde Z}^*\mb W\mb\Psi_{\widetilde Z},
    \qquad
    \mb B = \mb\Psi_Z^*\mb W\mb\Psi_{\widetilde Z},
    \qquad
    \mb G = \mb\Psi_Z^*\mb W\mb\Psi_Z .
\]
We use
\begin{equation*}
    \operatorname{res}(\lambda,\mb g)
    =
    \left(
    \frac{
        \mb g^*
        \left(
        \mb A
        -
        \lambda \mb B^*
        -
        \bar\lambda \mb B
        +
        |\lambda|^2 \mb G
        \right)
        \mb g
    }{
        \mb g^*\mb G\mb g
    }
    \right)^{1/2},
\end{equation*}
which is a data-driven approximation of
\[
    \frac{\|\mathcal K g-\lambda g\|}{\|g\|}
\]
\cite{colbrook2023residual,colbrook2024rigorous}. The top row of
\cref{fig:pendulum spectra} shows the computed eigenvalues, colored by this
residual. EDMD, although using the same dictionary as DeepMDMD, produces
spurious eigenvalues inside the unit disk. This ablation shows that the
multiplicative constraint is important independently of dictionary quality.
MDMD places its nonzero eigenvalues on $\mathbb T$, but they concentrate on
low-order cyclic subgroups and leave much of the continuous spectrum unresolved.
DeepMDMD also respects the unit circle, but resolves higher-order subgroups,
giving denser coverage of the continuous spectrum.

To quantify this coverage, we exploit the multiplicative structure in
\cref{closed_eigs}. Products of principal eigenfunctions generate further
eigenfunctions \cite{monfared2026algebra,mohr2016koopman}, and taking
logarithms turns products into sums. Thus, wherever
$\phi_1(\mb x)\phi_2(\mb x)\ne0$,
\begin{equation*}
    \log \bigl| \phi_1^n(\mb x)\phi_2^m(\mb x) \bigr|
    =
    n\log|\phi_1(\mb x)|
    +
    m\log|\phi_2(\mb x)| .
\end{equation*}
Eigenfunctions generated multiplicatively from existing ones therefore become
linearly dependent in log-modulus coordinates. This motivates the sample
matrix
\[
    \mb R
    =
    \bigl[
        \log|\phi_1|
        \ \cdots\
        \log|\phi_N|
        \bigr],
\]
whose numerical rank gives an empirical measure of how many algebraically
independent, or principal, eigenfunctions have been resolved. The bottom row
of \cref{fig:pendulum spectra} plots the singular values of $\mb R$ for
DeepMDMD, MDMD, and EDMD. DeepMDMD has substantially slower singular-value
decay than MDMD, indicating that it resolves more principal directions and
therefore gives a richer approximation of the pendulum's continuous spectrum.

\begin{figure}[th!]
    \centering
    \includegraphics[width=1\textwidth]{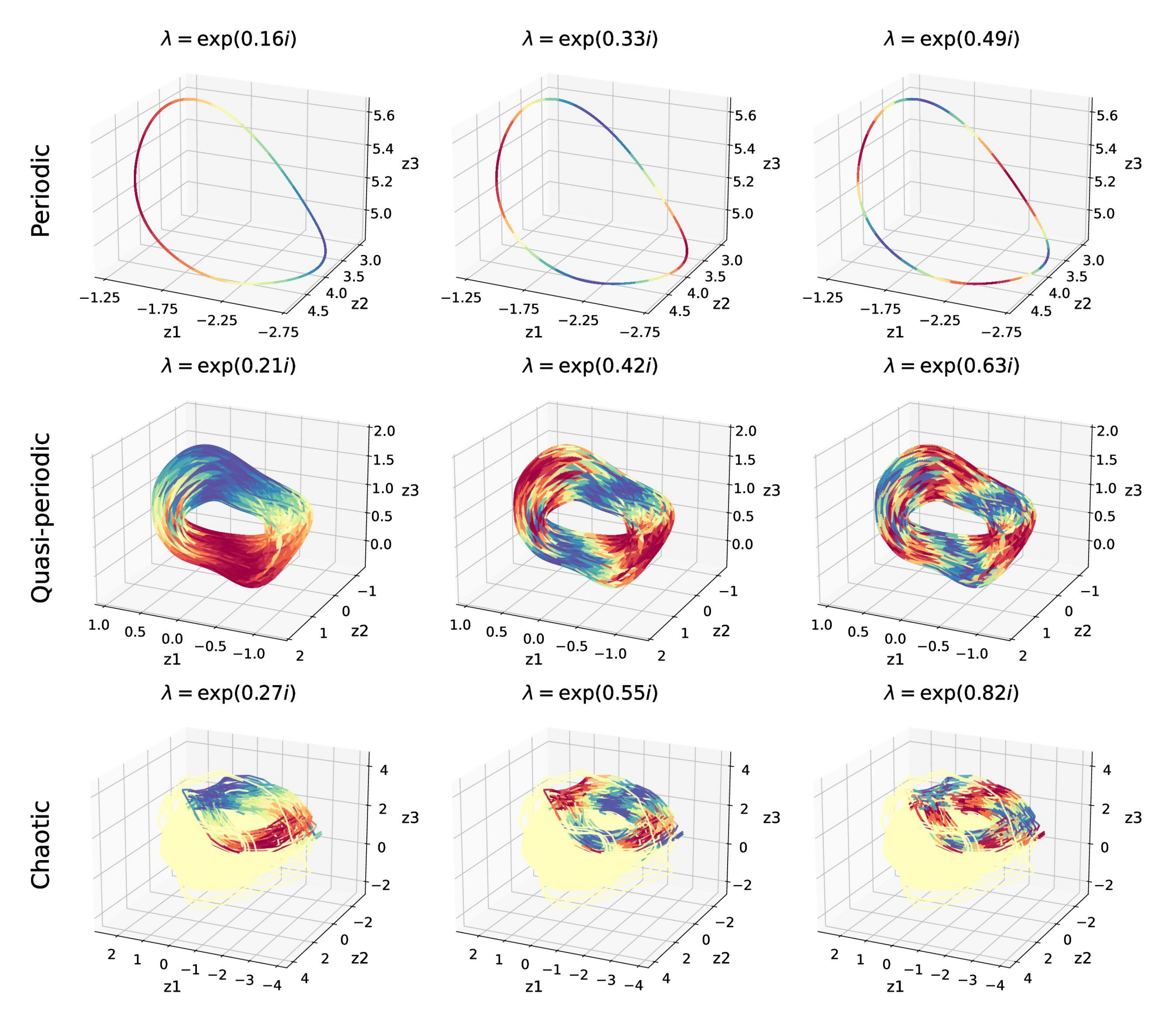}
    \caption{Koopman eigenfunctions of the Lorenz-$96$ system at successive bifurcations ($f=2.0,3.5,4.2$), visualized on the DeepMDMD latent space. As $f$ increases, the latent embeddings develop increasingly complex structure, reflecting the transition from periodic to chaotic dynamics.}
    \label{fig:lorenz96}
\end{figure}
\begin{figure}[t]
    \centering    \includegraphics[width=1\linewidth]{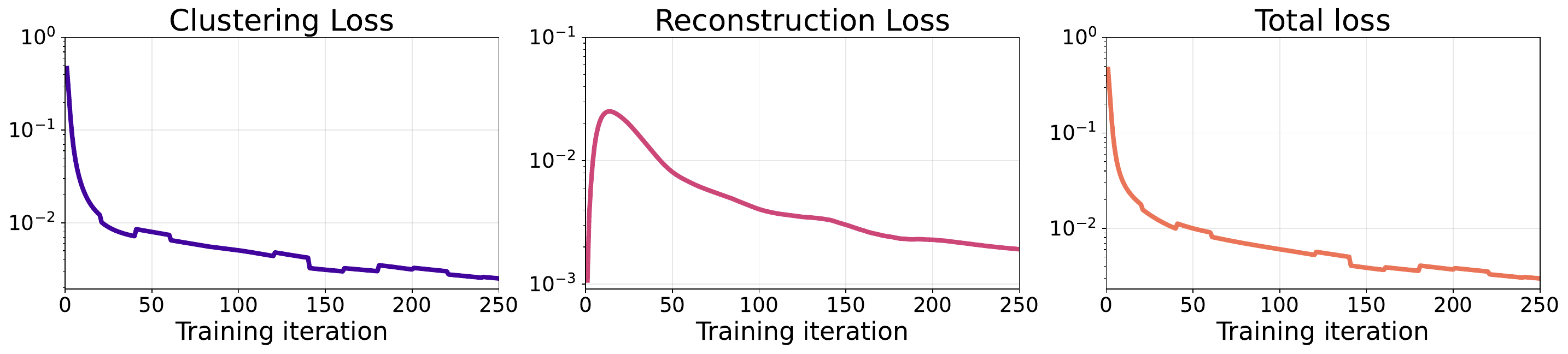}
    \caption{DeepMDMD training losses for the cylinder wake following the autoencoder warm-up, across alternating operator and dictionary update steps. The operator update is performed every $T_{\mathrm{update}} = 20$ gradient steps and the reconstruction loss is weighted by $\lambda = 0.25$.}
    \label{fig:loss_history}
\end{figure}

\subsection{Lorenz--96 attractors}

We next test DeepMDMD on the $d$-dimensional Lorenz--96 system
\cite{lorenz2017deterministic}, focusing on its ability to learn a
dynamically organized latent space and the associated Koopman spectral
structure. The equations are
\begin{equation*}
    \frac{dx_i}{dt}
    =
    (x_{i+1}-x_{i-2})x_{i-1}
    -
    x_i
    +
    f,
    \qquad i=1,\dots,d,
\end{equation*}
with indices interpreted modulo $d$. We select $d=9$ and consider the forcing values
$f\in\{2.0,3.5,4.2\}$, corresponding respectively to periodic, quasiperiodic,
and chaotic dynamics on the attractor \cite{van2018travelling}. From a random
initial condition, we integrate with a fourth-order Runge--Kutta method and
time step $\Delta t=0.1$, discarding the first $10^3$ snapshots as transient.
This gives $M=9\times 10^3$ training snapshot pairs.

For each value of $f$, we train DeepMDMD with $N=1000$ basis functions, latent
dimension $k=3$, and reconstruction weight $\lambda=0.25$. The encoder
$\mb E_\theta$ maps the nine-dimensional snapshots into
$\mathcal Z\subset\mathbb R^3$, where the MDMD partition and Koopman matrix
are learned. In \cref{fig:lorenz96} we show the resulting latent spaces
together with selected Koopman eigenfunctions. Each latent point
$\mb z^{(m)}$ is colored by the eigenvector entry associated with its assigned
cluster. As the forcing increases, the learned latent geometry changes from a
closed curve to a torus-like set and then to a chaotic attractor, consistent
with the bifurcation structure of the Lorenz--96 dynamics.

\subsection{Noisy cylinder wake}

We next consider a high-dimensional example: low-Reynolds-number flow past a
circular cylinder. The data are generated as in \cite{colbrook2024another},
using an incompressible two-dimensional lattice--Boltzmann solver
\cite{jozsa2016validation, szHoke2017performance}. Each snapshot is the
vorticity field on an $800\times 200$ grid. After masking the cylinder
boundary, the state dimension is $d=158,624$. The snapshots are standardized
before training. At $\mathrm{Re}=100$ the flow is periodic on the attractor,
and the Koopman spectrum is pure point. We use the first $M=80$ snapshot pairs
for training and the next $1000$ snapshots for testing.

\begin{figure}[t]
    \centering
    \includegraphics[width=0.85\linewidth]{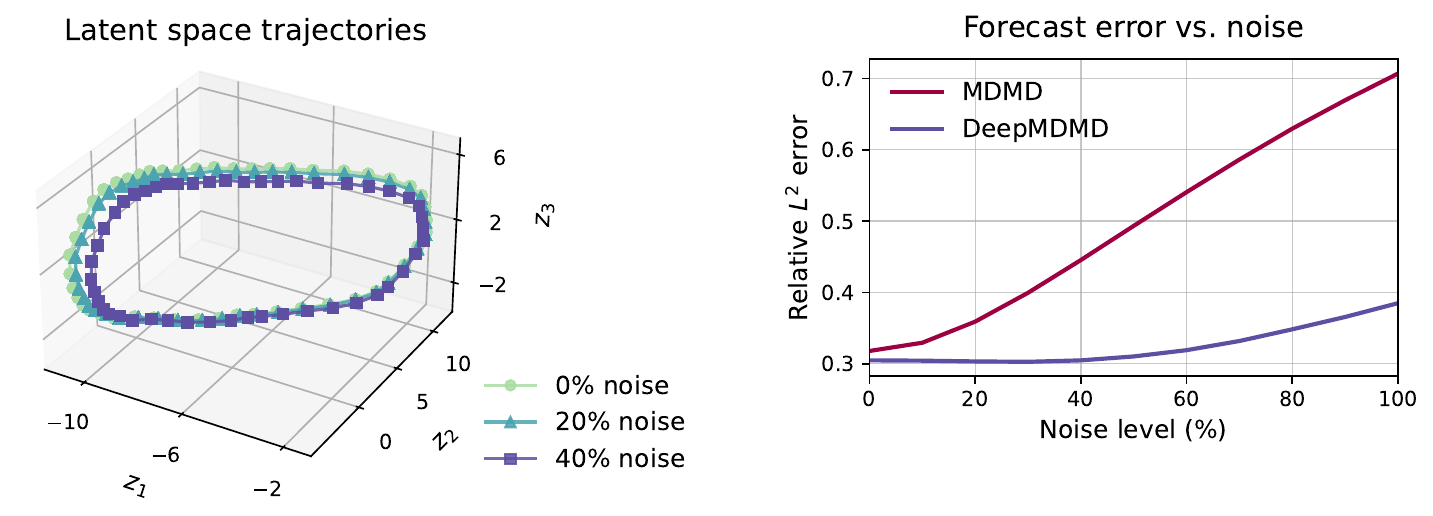}
    \caption{Cylinder wake with noise added after training. Left: DeepMDMD
        latent-space forecasts at different noise levels. Right: mean relative
        $L^2$ forecast error over the test window for DeepMDMD and MDMD as a
        function of noise level.}
    \label{fig:cylinder latents}
\end{figure}

\begin{figure}[t]
    \centering
    \includegraphics[width=0.75\linewidth]{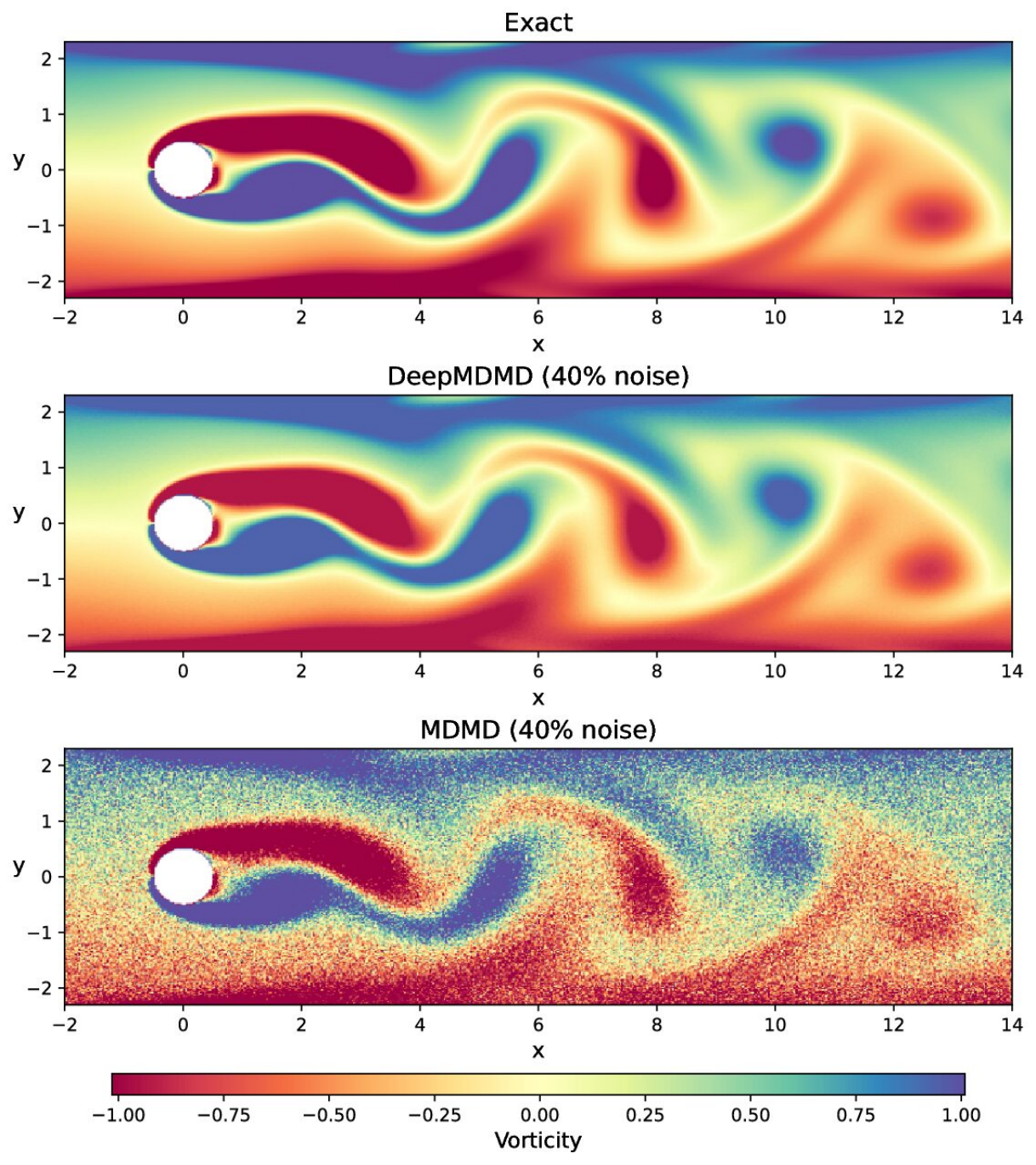}
    \caption{Cylinder wake. One-step forecasts from DeepMDMD and MDMD with
        $40\%$ Gaussian noise added after training.}
    \label{fig:cylinder predictions}
\end{figure}

We train DeepMDMD with dictionary size $N=80$, latent dimension $k=3$, and
reconstruction weight $\lambda=0.25$. The reconstruction term keeps the decoder
active during fine-tuning. In \cref{fig:loss_history} we show the DeepMDMD
training losses after autoencoder pretraining, illustrating stable convergence
of the alternating scheme. As a baseline, we fit MDMD with the same dictionary
size, using indicator functions obtained by $k$-means++ clustering of the
leading $k=3$ POD coordinates.

To test robustness, we add Gaussian noise to the snapshots after training. This
models a deployment setting in which the learned operator is fixed but the
measurements are corrupted. DeepMDMD forecasts are computed in latent space
using \cref{alg:DeepMDMD_forecast}, whereas MDMD forecasts use the
state-space KMD expansion \eqref{KMD 1}.

The results are summarized in \cref{fig:cylinder latents}. The left panel
shows DeepMDMD latent trajectories over the full $1000$-step forecast window.
Even with noisy inputs, the trajectories remain close to the intrinsic limit
cycle. The latent bottleneck filters components that are not represented by
the learned low-dimensional manifold, reducing the effect of off-manifold
noise on the rollout. The right panel shows the mean relative state-space
$L^2$ error over the forecast window. After decoding, DeepMDMD gives
substantially lower errors than MDMD, which operates directly in the
high-dimensional state space and has no comparable denoising mechanism.
\Cref{fig:cylinder predictions} shows representative one-step forecasts at
$40\%$ Gaussian noise, confirming the same behavior visually.

\begin{figure}[t]
    \centering
    \includegraphics[width=1\textwidth]{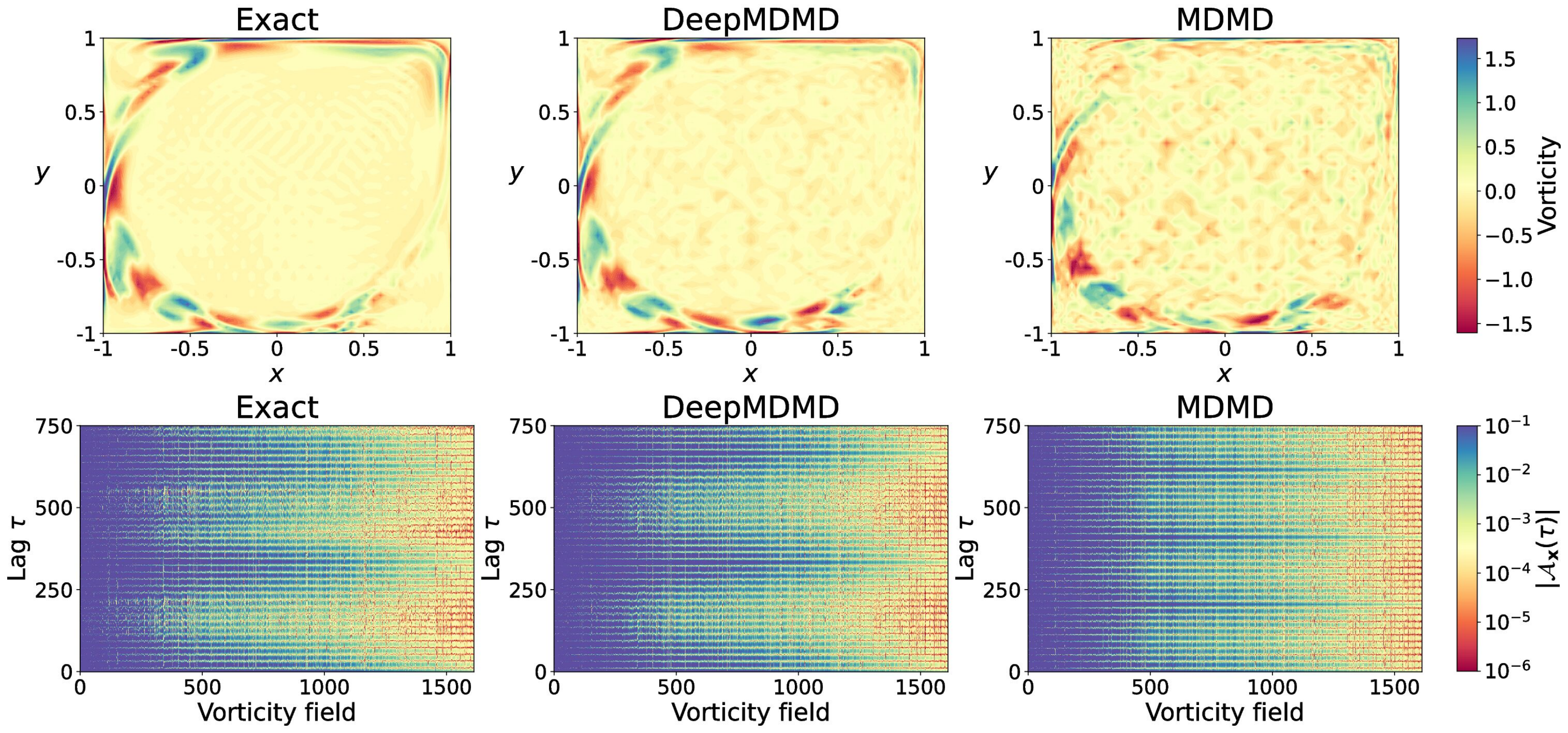}
    \caption{Noisy cavity flow experiment. Top: vorticity profiles predicted at timestep $t=100$. Bottom: absolute autocorrelation $|\mathcal{A}_{\mb x}(\tau)|$ computed via \eqref{autocorr} across lags $\tau=1,\dots,750$, displayed on a log color scale. Only spatial points whose time-averaged absolute vorticity exceeds a threshold of $\epsilon=0.25$ are retained.}
    \label{fig:autocorrelations}
\end{figure}

\subsection{Noisy lid-driven cavity}

Finally, we consider the two-dimensional lid-driven cavity. The system consists
of an incompressible viscous fluid in a cavity whose upper lid moves tangentially.
The data are taken from \cite{arbabi2017study} and consist of vorticity
snapshots on a $65\times 65$ grid, giving state dimension $d=4225$. The
Reynolds number is $\mathrm{Re}=20,000$, a regime with mixed Koopman spectral
content, including both discrete and continuous components. The snapshots are
mean-subtracted before training. The first $M=500$ snapshot pairs are corrupted
with $40\%$ Gaussian noise and used for training; the next $1000$ clean
snapshots are reserved for testing.
To assess whether the model captures the flow statistics, we compute
autocorrelations of the vorticity field. For an observable
$g\in L^2(\mathcal X,\omega)$, define
\begin{equation}
    \mathcal A_g(\tau)
    =
    \langle \mathcal K^\tau g, g\rangle,
    \qquad \tau\ge 1 .
    \label{autocorr}
\end{equation}
For a unitary Koopman operator, this sequence gives the trigonometric moments
of the spectral measure associated with $g$, and hence provides a useful
signature of the Koopman spectrum \cite{korda2020data}. We take
$g(\mb x)=\mb x$ and estimate \eqref{autocorr} from the forecast trajectory
using Birkhoff averages, applying the computation componentwise and summing
over selected spatial degrees of freedom. To focus on dynamically active
regions, we retain only grid points whose time-averaged absolute vorticity
exceeds $\epsilon=0.25$.
In \cref{fig:autocorrelations} we plot $|\mathcal A_{\mb x}(\tau)|$ for the
ground truth, DeepMDMD, and MDMD. DeepMDMD tracks the ground truth over most
time lags, suggesting that it captures the dominant spectral statistics of the
flow. MDMD, by contrast, does not reproduce the observed autocorrelation
structure.

\section{Conclusion}\label{conclusion}

We have introduced Deep Embedded Multiplicative Dynamic Mode Decomposition, a method for learning structure-preserving Koopman approximations
in a latent space. The central idea is to move the MDMD partition from the
ambient state space to a learned latent representation, and then to
shape this representation by the dynamics. The resulting Koopman matrix is not
merely fitted to data: it is constrained exactly to preserve the multiplicative
structure of the Koopman operator. Thus DeepMDMD combines the flexibility of
neural representations with the stability and spectral structure of MDMD.

The numerical experiments show the value of both ingredients. On the nonlinear
pendulum, DeepMDMD learns partitions aligned with invariant energy levels,
leading to compact dictionaries, reduced spectral pollution, and a richer
approximation of the continuous spectrum than standard MDMD. On the Lorenz--96
system, the learned latent spaces recover the transition from periodic to
quasiperiodic and chaotic dynamics. On the cylinder wake and lid-driven cavity,
DeepMDMD produces stable latent-space rollouts and accurate flow statistics in
high-dimensional, noisy settings. These examples suggest that the method is
most useful when geometry alone is a poor guide to the dynamics: the partition
must be learned, but the Koopman approximation must still respect structure.

The main cost of DeepMDMD is training. This cost is front-loaded, however: once
the latent space, partition, and Koopman matrix have been learned, forecasting
takes place in the latent space and decoding is needed only for output. The
method is therefore especially attractive for high-dimensional systems, where
full-state Koopman mode decompositions are expensive and sensitive to noise. Several directions remain open. One is to choose the latent dimension
automatically, for example by estimating the intrinsic dimension required by
the dynamics \cite{zeng2024autoencoders}. Another is to combine deep latent
representations with other structure-preserving Koopman approximations. More
broadly, DeepMDMD points to a useful principle for data-driven Koopman
learning: learn the coordinates, but constrain the operator.

\section*{Acknowledgments}
This work was supported by the UK Engineering and Physical Sciences Research Council (EPSRC) grant EP/Y034767/1.

\bibliographystyle{jabbrv_plain}

\bibliography{references}

\end{document}